\documentclass{article}

\usepackage[nonatbib,final]{nips_2017}

\usepackage[utf8]{inputenc} 
\usepackage[T1]{fontenc}    
\usepackage{hyperref}       
\usepackage{url}            
\usepackage{booktabs}       
\usepackage{amsfonts}       
\usepackage{nicefrac}       
\usepackage{microtype}      
\usepackage{filecontents}
\usepackage{amsmath}

\usepackage{graphicx}
\usepackage{wrapfig}

\usepackage[numbers]{natbib}

\newcommand{\cF}{ {\cal F} }
\newcommand{\cD}{ {\cal D} }
\newcommand{\cL}{ {\cal L} }
\newcommand{\vU}{ \mathbf{U} }
\newcommand{\XX}{ {\cal X} }
\newcommand{\NN}{ {\cal N} }
\newcommand{\RR}{ \mathbb{R} }

\newcommand{\EE}{ \mathbb{E} }

\title{Structured Variational Inference for Coupled Gaussian Processes}

\author{
  Adam Vincent \\
  Gatsby Computational Neuroscience Unit\\
  University College London\\
  London, W1T 4JG \\
  \texttt{vincenta@gatsby.ucl.ac.uk} \\
}

\begin{document}

\maketitle

\begin{abstract}
  Sparse variational approximations allow for principled and scalable inference in Gaussian Process (GP) models.
  In settings where several GPs are part of the generative model, these GPs are a posteriori coupled. 
  For many applications such as regression where predictive accuracy is the quantity of interest, this coupling is not crucial. Howewer if one is interested in posterior uncertainty, it cannot be ignored. 
  A key element of variational inference schemes is the choice of the approximate posterior parameterization.
  When the number of latent variables is large, mean field (MF) methods provide fast and  accurate posterior means while more structured posterior lead to inference algorithm of greater computational complexity. Here, we extend previous sparse GP approximations and propose a novel parameterization of variational posteriors in the multi-GPs setting allowing for fast and scalable inference capturing posterior dependencies.
  
\end{abstract}

\section{Introduction}

Gaussian Processes (GPs) are powerful nonparametric distributions over continuous functions which can be used for both supervised and unsupervised learning problems \cite{rasmussen2006gaussian}.
Here, we propose an extension of the variational pseudo-point GP (VFE) approximation \cite{titsias2009variational, bauer2016understanding} to the case of regression models composed of multiple GPs. Such models include Generalized Additive Models (GAM) which are a commonly used non-linear extension of generalized linear models when interpretability of models is required \cite{hastie1990generalized}. In these models, inference leads to posterior dependencies across GPs. For this reason we will refer to this setting as Coupled GPs (CGPs).

The VFE approximation provides state-of-the art performance for GP regression and provides approximations to the posterior distributions in the form of a GP.
This approach has been successfully extended to the CGPs setting using a factorized (mean-field) approximation of the posterior across GPs \cite{saul2016chained,adam2016scalable}. However, it suffers from the known variance underestimation of mean-field approximations and therefore can lead to poor predictions or can bias learning \cite{turner2011two}. Furthermore the mean-field approaches provide no information about how the uncertainties across GPs interact.

Several general approaches have been proposed to alleviate this problem of mean field approximations \cite{giordano2015linear,tran2015copula, hoffman2015stochastic}, however, they do not readily apply to the CGPs setting or cannot leverage the sparse approximation.

This paper introduces a new posterior approximation tailored to the CGPs setting that explicitely captures posterior dependencies and allows for scalable inference. We now briefly review the VFE approach to GP regression and develop the proposed Variationally Coupled GPs (VCGPs) approach.

\section{Regression with multiple GPs}

We are interested in the broad class of regression models with predictors of the form 
$\rho\left(x\right)=\Phi\left(f_{1}\left(x\right),...,f_{C}\left(x\right)\right)$ where $f_c$ are functions from $\XX_c \to \RR$. We consider models with factorized likelihood of the form $p\left(y|X\right)=\prod_{i}p\left(y_{i}|\rho\left(x_{i}\right)\right)$. The specific form of the likelihood is arbitrary.  $\Phi$ is an arbitrary function from $\RR^C \to \RR$ and is assumed to be known. This model class generalizes the classical GP regression setting as well as GLMs and GAMs.

 We are interested in inferring the functions $f_{1..C}$ from data.
Without further assumptions, this is an ill-posed problem. Here, we are interested in the case where independent GPs are used as priors over the different latent functions and we wish to perform Bayesian inference. For most choices of kernels, likelihood and function $\Phi$, latents are coupled a posteriori.

We denote $\cF=f_{1},...,f_{C}$ such that $p\left( \cF \right)$ constitutes the joint distribution over the processes. 
$\cF \left(x\right)= [ f_{1}\left(x\right),...,f_{C}\left(x\right)]$
 is the vector of function evaluations at $x$. We are interested in computing the joint posterior $p( \cF |X,Y)$.

\section{Pseudo-point approximations in multiple GPs regression}
 
\subsection{Variational lower bound to the log evidence}

The classical variational lower bound is  
\begin{equation} \label{eq:bound1}
\log \,p(y|X) \geq E_{q\left(\cF \right)}\log\:\frac{p\left(y|\cF,X\right)p\left(\cF\right)}{q\left(\cF\right)} = \cL(q)
\end{equation}

where $q$ is the variational posterior approximation.

Following \cite{saul2016chained,adam2016scalable} we introduce for each GP indexed $c$ some 'inducing points' $Z_{c}=[z_{c}^{(1)},...,z_{c}^{(m)}]\in \XX_{c}^{m}$. We note $\vU_{c}$ the vector of associated function evaluations $\vU_{c}=[u_{c}^{(1)},...,u_{c}^{(m)}]=[f_{c}(z_{c}^{(1)}),...,f_{c}(z_{c}^{(m)})]$. We also note $\vU=\left[\vU_{1},...,\vU_{c}\right]$ the stacked vector.

\subsection{Parameterization of the coupled variational posterior}

We then choose the following form for our posterior approximation $q\left(\cF\right)=q\left(\vU\right)\prod_{c}p(f_{c\neg\vU_{c}}|\vU_{c})$. This intuitively means that the posterior is compressed into the lower dimensional $q(\vU)$ since the full posterior $q\left(\cF\right)$ is built for each latent by interpolating from $q(\vU)$ using the prior only.

This choice leads to a simplification of the lower bound (\ref{eq:bound1}) as
\begin{align}
 \label{eq:bound2}
 \cL(q) &= E_{q(\cF)} \log \: \frac{p(y|\cF,X)p\left(\vU\right)\prod_{c}p(f_{c\neg\vU_{c}}|\vU_{c})}{q\left(\vU\right)\prod_{c}p(f_{c\neg\vU_{c}}|\vU_{c})} \\
 &= E_{q(\cF)} \log \: p(y|\cF,X) - KL[q\left(\vU\right)||p\left(\vU\right)]
\end{align}
where KL denotes the Kullback-Leibler divergence between distributions. 

We also restrict ourselves to the case of factorizing likelihoods which simplifies the expectations terms in the bound
\begin{equation}
 \label{eq:bound3}
 \cL(q) = \sum_i E_{q(\cF^{(i)})} \log \: p(y^{(i)}|\cF^{(i)},x^{(i)}) - KL[q\left(\vU\right)||p\left(\vU\right)]
\end{equation}
Saul et al \cite{saul2016chained} considered the mean field case $q\left(\vU\right)=\prod_{c}q\left(\vU_{c}\right)
$ with each factor parameterized as a multivariate Gaussian distribution \NN($\mu_{\vU_c}$ ,$\Sigma_{\vU_c})$. This approach does not capture posterior dependencies across GPs. Instead, we here directly parameterize $q\left(\vU\right)$ a multivariate Normal distribution \NN($\mu_{\vU}$ ,$\Sigma_{\vU})$ and show how it only induces small additional complexity and how methods to train VFE for single GPs readily apply to our setting \cite{hensman2015scalable}.

\section{Optimizing the variational lower bound}

The Kullback divergence term in equation (\ref{eq:bound3}) is between two multivariate Gaussian distributions. Its expression and derivatives are available in closed form.
The expectation terms in equation (\ref{eq:bound3}) are intractable in most cases and need to be approximated.

\subsection{Stochastic optimization}

We approximate the expectation terms in (\ref{eq:bound3}) with Monte Carlo samples from the variational posterior for each data point.
For each expectation $E_{q(v)}f(v)$ under $v \sim \NN(\mu,\Sigma=LL^T)$, we reparameterize the Gaussian distribution as $v=L \epsilon + \mu$ with $\epsilon\sim \NN(0,I)$. Using a sampling procedure in terms of isotropic Gaussians, we can compute unbiased gradients of the bound with respect to the variational parameters \cite{salimbeni2017doubly}.
Since the bound is expressed as a sum of terms depending on each data-point separately, we can scale the methods to large datasets by sub-sampling the data.

\subsection{Complexity}

Given $C$ latents, with $M$ inducing points each, this leads to a vector $\vU$ of size $MC$ and hence a number of varational parameters that is $O(M^2 C^2)$.
The evaluation of the lower bound requires the computation of the joint $q(\cF(x)) =  \int d\vU  \:p(\cF(x)|\vU)q(\vU)$ for each data point which is $O(C^4 M^2)$, leading to a cost of $O(N C^4 M^2)$ that is linear in $N$. Finally, computing the KL divergence of the bound requires $O(C^3 M^3)$.

\section{Experiment: Conjugate Additive Regression}

We demonstrate the performance of our VCGP approach in a preliminary experiment. We show that we can achieve tighter bounds to the marginal evidence in the conjugate addtive setting where both the marginal evidence and the full coupled posterior are available in closed form.\\
This experiment was implemented using GPflow \cite{GPflow2017}

\subsection{Setting}

As a toy example, we consider the case of GP additive regression \citep{adam2016scalable} with Gaussian observation noise and $N=500$ data points. Ground truth functions are set to $f_1(x) = \sin(x)^3$ and $f_2(x) = \cos(3 x)$. Covariates are sampled as $x_1,x_2 \sim Uniform[-3,3]$. The observed output is $y^i = f_1(x^i_1) + f_2(x^i_2) + e^i$ with 
$e^i \sim \NN (0,\sigma^2)$ and $\sigma=0.5$. Priors for both latents are set to be GPs with $0$-mean and Gaussian kernels $k_c(x,x')=s_c^2 \exp \left[ \frac{(x-x')^2}{2 l_c^2} \right]$. Parameters  $\sigma, s_1,s_2,l_1,l_2$ are set to maximize the marginal evidence $p(y|x)$ that is available in closed form, and held fixed throughout the
optimization of the MF and VCGP models. Variational inference is then performed in these two models. In both cases, each latent is approximated using $10$ and $30$ inducing points whose position is optimized.

\subsection{Results}

\begin{table}[t]
  \caption{Results - Conjugate Additive Regression}
  \label{res-table}
  \centering
  \begin{tabular}{llllll}
    \toprule

Model & RMSE &  $\sqrt{\EE_\cD[\sigma^2_{post}(\sum f)}]$ & $\EE_\cD[\rho_{post}(f_1,f_2)]$ & $\sqrt{\EE_\cD[\sigma^2_{post}(\Delta f)}]$ &  $-\cL(q)$ \\
\midrule
Exact    & 0.4675 &                0.0967 &               -0.9306 &               0.509  &                 381.924  \\
VCGP[10] & 0.4698 &                0.1237 &               -0.892  &               0.5109 &                 388.667  \\
VCGP[30] & 0.4678 &                0.1036 &               -0.9265 &               0.5286 &                 383.714  \\
MF[10]   & 0.469  &                0.1331 &                0      &               0.1331 &                 398.376  \\
MF[30]   & 0.468  &                0.125  &                0      &               0.125  &                 409.01   \\
\bottomrule
  \end{tabular}
\end{table}

All results are summarized in Table~\ref{res-table}. Both the exact, MF and SVGP models perform equally well in terms of predictive accuracy (RMSE) with a small tendency to overestimate the posterior predictive uncertainty on the sum of the latents ($\Sigma f=f_1+f_2$). 

\begin{figure}[h]
  \centering
  \includegraphics[width=12cm]{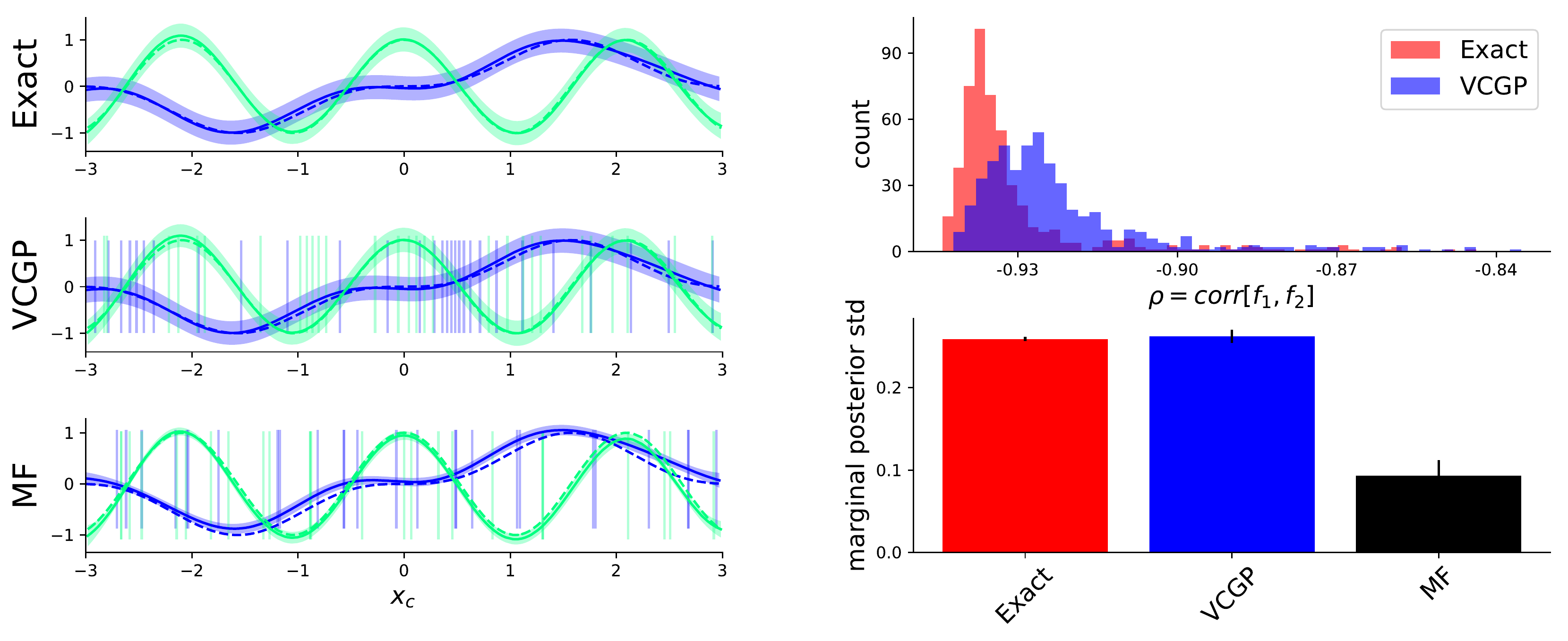}
  \caption{Output of inference for exact, VCGP[30] and MF[30] model.  \emph{Left:} Dashed line are the true latents, tubes represent the marginal posterior GP for each individual source. For the variational algorithms, vertical bars show the positions of the pseudo inputs associated to the inducing points. \emph{Right, Top:} Histogram of posterior correlation $\rho(f_1,f_2) = \frac{cov(f_1,f_2)}{\sqrt{var(f_1)var(f_2)}}$ across the two latents for the exact and the VCGP model. \emph{Right, Bottom:} Mean and variance of the marginal posterior standard deviations for all models.}
\label{fig:exp1_latent}
\end{figure}

However, only the SVGP model can provide accurate marginal variance predictions for each latent as is shown in Figure~\ref{fig:exp1_latent}(left)(bottom-right). It is also the only model to accurately predict the posterior correlation across latents and can thus correctly predict the posterior uncertainty on the difference between the latents ($\Delta f =f_1-f_2$). A histogram of these posterior correlations $\rho(f_1,f_2)$ for the SVGP model is shown in Figure~\ref{fig:exp1_latent}(top-right).

Furthermore, our model provides a tighter bound to the log-evidence (written $\cL(q)$ for the exact model in Table~\ref{res-table} for simplicity). This means that our coupled variational posterior $q_{\text{coupled}}$ is closer the true posterior $p$ than MF posterior $q_{\text{MF}}$ in the sense that $KL(q_{\text{coupled}}|p)<KL(q_{\text{MF}}|p)$.

Similarly to the standard VFE approach to GP regression \cite{bauer2016understanding}, the performance of our VCGP model improves as we increase the number of inducing points.

\section{Related work}

Variational inference for the CGP setting has so far only used the mean-field approximation as described in \cite{saul2016chained}. 
When posterior dependencies are a quantity of interest, a natural approach is to increase the complexity of the variational posterior to capture these dependencies. This often results in a prohibitive increase in the complexity of the inference. 

Different solutions have been proposed to tackle this problem.
A first approach in \cite{giordano2015linear} consists in a two step scheme where MF inference is \emph{assumed} to provide accurate posterior mean estimates. A perturbation analysis is then performed around the MF posterior means to provide second order (covariance) estimates.

A second approach consists in 'relaxing' the MF approximation by extending the variational posterior $q$ with addititional multiplicative terms capturing dependencies while keeping the computational complexity of the resulting inference scheme low \cite{tran2015copula,hoffman2015stochastic}.

Our approach fits in this second family of extensions of the the MF parameterization. It is taylored to the VFE approximation to GP models and leverages its sparsity to provide a fast and scalable inference algorithm.

\section{Conclusion}

We presented an extension of the VFE approximation to regression settings with mutliple GPs that takes advantage of its scalability and generality. Our approach goes beyond the classical mean-field approximation usually used and provides richer posteriors capturing dependencies between inferred latent functions. This approach is very general and provides an efficient way to perform inference in a class of models broadly used in statistics such that of GAMs.

\subsubsection*{Acknowledgments}

I would like to thank Lea Duncker, Eszter Vertes and Maneesh Sahani from the Gatsby Unit for helpful discussions.

\bibliographystyle{unsrt}

\end{document}